\documentclass[letterpaper, 10 pt, conference]{ieeeconf}  

\IEEEoverridecommandlockouts                              
\makeatletter
\def\bstctlcite{\@ifnextchar[{\@bstctlcite}{\@bstctlcite[@auxout]}}
\def\@bstctlcite[#1]#2{\@bsphack
  \@for\@citeb:=#2\do{%
    \edef\@citeb{\expandafter\@firstofone\@citeb}%
    \if@filesw\immediate\write\csname #1\endcsname{\string\citation{\@citeb}}\fi}%
  \@esphack}
\makeatother

\usepackage[vlined, ruled, linesnumbered, commentsnumbered]{algorithm2e}
\usepackage{amsmath}
\usepackage{cite}
\usepackage{color}
\usepackage{xcolor}
\definecolor{chred}{rgb}{0.8,0,0}
\definecolor{chgrey}{rgb}{0.5,0.5,0.5}
\usepackage{graphicx}
\usepackage{caption}
\usepackage{subcaption}
\usepackage{dblfloatfix}
\usepackage{eqlist}
\usepackage{txfonts}
\usepackage{url}
\usepackage{footmisc}
\usepackage{booktabs}
\usepackage{balance}

\usepackage{multirow}
\usepackage[para,online,flushleft]{threeparttable}
\DeclareMathOperator{\sgn}{sgn}

\title{\LARGE \bf Quickly Inserting Pegs into Uncertain Holes using Multi-view Images and Deep Network Trained on Synthetic Data}

\author{Joshua C. Triyonoputro$^{1}$, Weiwei Wan$^{1,2,*}$, Kensuke Harada$^{1,2}$%
\thanks{$^{1}${Graduate School of Engineering Science, Osaka University, Japan.}
$^{2}${National Inst. of AIST, Japan.} *{Correspondent author: Weiwei Wan,
}{\tt\small wan@sys.es.osaka-u.ac.jp}}
}

\begin{document}
\maketitle
\thispagestyle{empty}
\pagestyle{empty}

\begin{abstract}
This paper uses robots to assemble pegs into holes on surfaces with different colors and textures.
It especially targets at the problem of peg-in-hole assembly with initial position uncertainty.
Two in-hand cameras and a force-torque sensor are used to account for the position uncertainty.
A program sequence comprising learning-based visual servoing, spiral search, and impedance control
is implemented to perform the peg-in-hole task with feedback from the above
sensors. Contributions are mainly made in the learning-based visual servoing
of the sequence, where a deep neural network is trained with various sets of synthetic
data generated using the concept of domain randomization to predict where a hole is. 
In the experiments and analysis section, the network is analyzed and compared,
and a real-world robotic system to insert pegs to holes using the proposed method is implemented.
The results show that the implemented peg-in-hole assembly system can perform
successful peg-in-hole insertions on surfaces with various colors and textures. 
It can generally speed up the entire peg-in-hole process.
\end{abstract}

\section{Introduction} \label{introduction}

One goal in robotics is to automate product assembly. At present, most robotic 
assembly systems, such as the ones implemented in the production of cars, still
follow the basic principle of teaching and playback. The teaching and playback concept is
useful when the target objects are fixed. When variations exist, the usefulness
of teaching and playback principle is limited. This leads people to use automatic
motion planning to assemble a diverse range of products with variations.

An important issue of automatic motion planning is the accumulated position errors. The
goal pose after execution could be very different from the goal
pose in the simulation after motion planning. People usually use
visual detection or scanning search using force sensors to locate the hole and 
avoid the position errors. However, both methods have shortages: Visual detection requires
mild color, texture, and reflection, etc.; Scanning search using force sensors is slow.

\begin{figure}[!htbp]
    \centering
    \includegraphics[width=0.8\columnwidth]{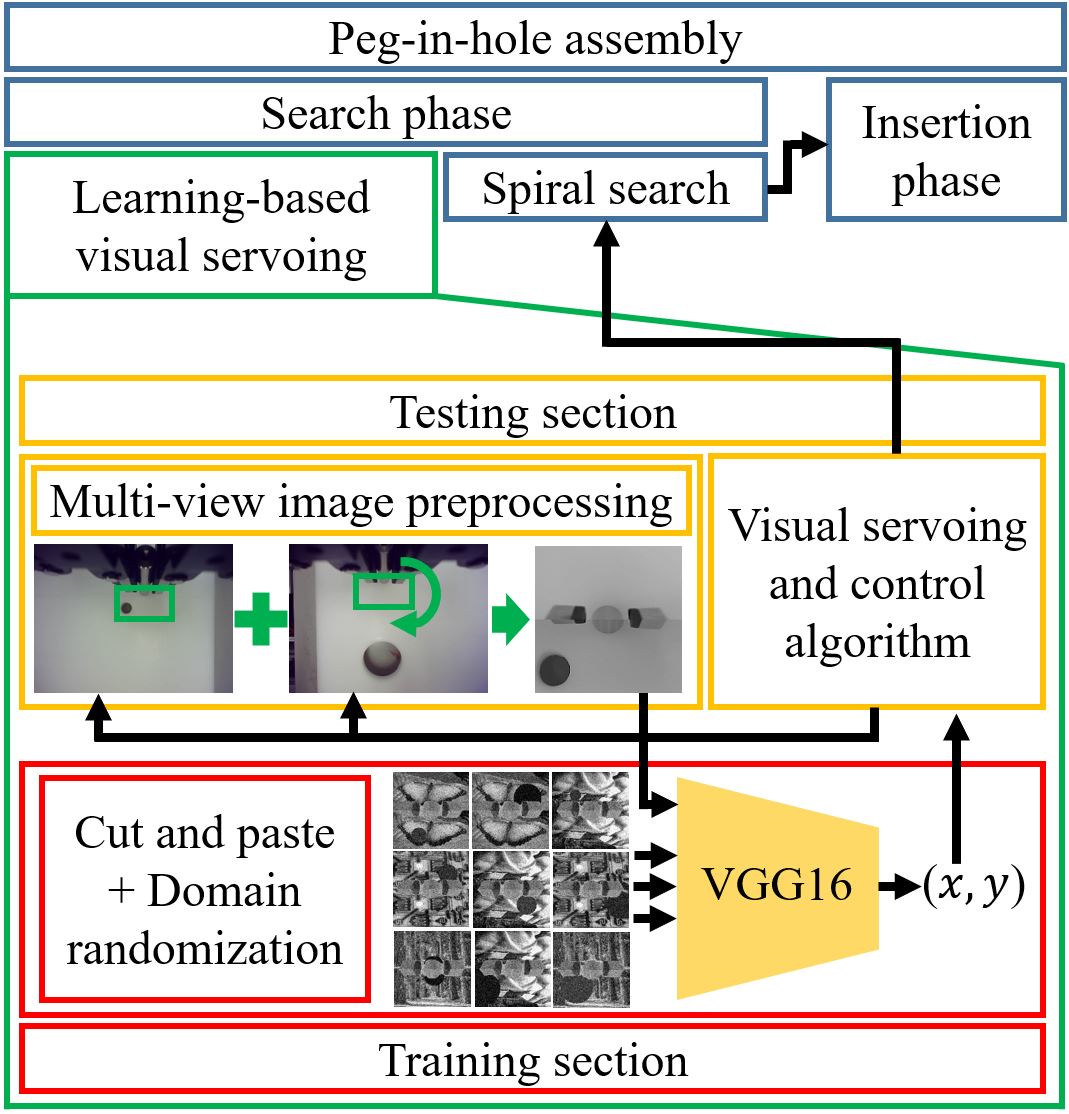}
    \caption{
    The workflow of the proposed peg-in-hole assembly system. The system uses learning-based visual servoing to quickly
    move the peg closer to the hole, uses spiral search to precisely align the peg and the hole, and uses
    impedance control to fully insert the peg into the hole. The learning-based visual servoing is our main contribution.}
    \label{fig:first_image}
\end{figure}

This leads us to study how to quickly assemble pegs into uncertain holes on 
surfaces with different colors and textures.
We develop a peg-in-hole assembly system that uses learning-based visual servoing to quickly
move the peg closer to the hole, uses spiral search to precisely align the peg and the hole, and uses
impedance control to fully insert the peg into the hole.
Fig.\ref{fig:first_image} shows the outline of the developed peg-in-hole program. 
The search phase comprises the learning-based servoing and spiral search.
The insertion phase comprises the impedance control.

Specifically, our main contribution is the learning-based visual servoing. We use
synthesized data to train a deep neural network to predict the position of a hole,
and use iterative visual servoing to iteratively moves a peg towards the hole.

Various experiments and analysis using both simulation and real-world experiments 
are performed to (1) analyze the performance of the learning-based visual servoing
against uncertain holes on surfaces with different colors and textures, and (2)
compare the efficiency of executions under different initial hole positions.
The results show that the proposed method is robust to various surface backgrounds and
can generally speed up the entire peg-in-hole process.

\section{Related Work}\label{related_work}

This paper focuses on the problem of peg-in-hole assembly using deep learning. Thus,
this section reviews the related work in peg-in-hole assembly and the applications
of deep learning in industrial robots.

\subsection{Peg-in-hole assembly}\label{peginholeassembly}
Peg-in-hole assembly refers to the task of inserting a peg to a hole. The task
generally has two phases -- the search phase and the insertion phase. 
The insertion phase refers to the phase when the peg is being inserted,
and it has been studied extensively. 
The search phase is the stage of finding a hole when position uncertainty exceeds
the clearance of a hole. It is less studied.

\subsubsection{Insertion phase} One of the earliest studies about the insertion phase is
Shirai and Inoue\cite{shirai1973guiding}, where they used visual feedback to
perform insertion. About a decade later, researchers shifted from the use of
visual feedback to the use of compliance to accommodate the motion of the end-effector
during insertion\cite{whitney1982quasi}\cite{mason1981compliance}\cite{balletti2012towards}\cite{zhang2017peg}. In the 1990s, 
the quasi-static contact analysis was used to guide the insertion\cite{nguyen1995fuzzy}\cite{kim1999active}\cite{su2012sensor}. 
The state-of-the-art method for insertion is impedance control\cite{broenink1996peg}\cite{cho2012strategy}. It is
widely used in many practical systems.

\subsubsection{Search phase}
The search phase is before the insertion and is used to align the peg and the hole. 
Below, related work about the search phase, sometimes followed by insertion, is reviewed.
The studies can generally be categorized by the types of sensors used: vision sensors, force sensors, or both.

The first category uses vision sensors. Yoshimi and Allen\cite{yoshimi1994active}
dealt with visual uncertainty for peg-in-hole by attaching a camera to the
end-effector and rotating the camera around the last axis of the robot.
Morel et al.\cite{morel1998impedance} employed 2D visual servoing (search phase)
followed by force control (insertion phase) to successfully performed peg-in-hole
assembly with large initial offsets. Huang et al.\cite{huang2013fast} used
high-speed cameras to align a peg to a hole. More recently, the visual coaxial
system was used to perform precise alignment in peg-in-hole assembly\cite{nagarajan2016vision}\cite{yang2018coaxial}.

The second category uses force-torque sensors. Newman et al.\cite{newman2001interpretation}
proposed the use of force/torque maps to guide the robot to the hole. Sharma et al.\cite{sharma2013intelligent}
generalized the work of Newman et al.\cite{newman2001interpretation} to tilted pegs.
Chhatpar and Branicky\cite{chhatpar2001search} explored various blind search
methods such as tilting and covering the search space using paths like spiral path (a.k.a. spiral search).
Spiral search and its variants were also discussed in \cite{jasim2014position}\cite{marvel2018multi}.
The problem of spiral search is that it is time consuming, given that the robot
just blindly searches for the hole. Tilting, often considered as a method intuitive
to human, was also explored in several studies\cite{marvel2018multi}\cite{park2013intuitive}\cite{jasim2014contact}\cite{abdullah2015approach}.
The limitation of tilting is that it assumes that the initial offset is small. 

More recently, studies such as \cite{nguyen2019probabilistic} used a combination of visual sensors
and force-torque sensors to track the uncertainties of object poses and sped up the search process.
This paper similarly employs both visual sensors and a force-torque sensor. Specifically,
we explore the use of a combination of visual servoing using two in-hand RGB cameras, 
followed by the spiral search using force sensors, to perform a peg-in-hole assembly.

\subsection{Deep learning in industrial robotics}
Deep learning has in the recent years gained prominence in robotics.
Some studies used real-world data to train deep neural networks. For example,
Pinto and Gupta\cite{pinto2016supersizing} collected 700 robot hours of data and 
used them to train a robot to grasp objects. Levine et al.\cite{levine2018learning}
similarly made robots perform bin-picking randomly for days, before finally using
the data obtained to train a deep network for bin-picking. Inoue et al.\cite{inoue2017deep}
proposed deep reinforcement networks for precise assembly tasks. Lee et al.\cite{lee2018making}
trained multimodal representations of contact-rich tasks and trained a robot
to perform peg-in-hole. Thomas et al.\cite{thomas2018learning} used CAD data to
help improve the performance of end-to-end learning for robotic assembly. Yang et al.\cite{yang2017repeatable} 
and Ochi et al.\cite{ochi2018deep} took data by performing teleoperations
and used them to make the robot learn specific motions.
De Magistris et al.\cite{de2018experimental} took labeled force-torque sensor
data to train a robot to perform multi-shape insertion.
Although the aforementioned studies showed the possibilities of using real-world data, 
developing such systems are difficult.
Collecting and labeling the real-world data is time-consuming and labor intensive.

For this reason, robotic searchers began to study training deep neural networks using synthesized data.
Dwibedi et al.\cite{dwibedi2017cut} cut and pasted pictures of objects on
random backgrounds to train deep neural networks for object recognition. 
Mahler et al.\cite{mahler2017dex} used synthetic data of depth images to train
robots for bin-picking. 
Unfortunately, the use of synthetic data is limited due to reality
gap\cite{jakobi1995noise}.

To overcome the reality gap, one method is transfer
learning \cite{pan2010survey}. Domain adaptation is an example of transfer learning,
where synthetic data and real-world data are both used\cite{csurka2017domain}\cite{zhang2017adversarial} \cite{bousmalis2018using}\cite{fang2018multi}.
Domain randomization highly promotes the use of synthesized data\cite{sadeghi2016cad2rl}\cite{tobin2017domain}. 
It suggests that synthetic data with randomization can
be helpful to allow transfer without the need for real-world data.
Using domain randomization in data synthesis were widely studied\cite{tremblay2018training}\cite{andrychowicz2018learning}\cite{prakash2018structured}\cite{sundermeyer2018implicit}.

The learning-based visual servoing of this work is based on domain randomization.
It is used to synthesize the training data for a deep neural network.
It is also used to synthesize various testing data sets to analyze the performance of
the neural network.

\section{Methods}\label{method}
This section gives a general explanation of the proposed peg-in-hole assembly method,
with a special focus on the learning-based visual servoing, and the synthetic data generation.

\subsection{Overview of the proposed peg-in-Hole assembly}\label{proposed_peg_in_hole_method}
\begin{figure}[!htbp]
    \centering
    \includegraphics[width=0.95\columnwidth]{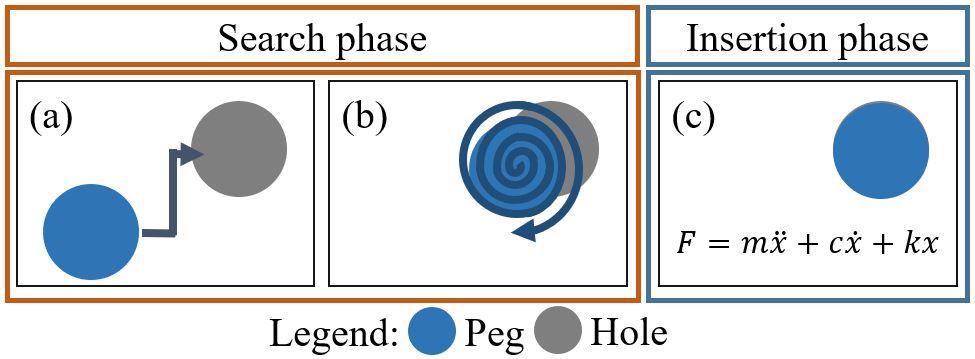}
    \caption{The proposed peg-in-hole assembly includes two phases: A search phase and an insertion phase.
    The search phase has two steps: 
    (a) Learning-based visual servoing and (b) Spiral search. They move and align the peg to the hole.
    The insertion phase uses (c) impedance control to insert the peg into the hole.}
    \label{fig:peg_in_hole_approach}
\end{figure}

The work flow of the proposed method is shown in Fig.\ref{fig:peg_in_hole_approach}.
It includes a search phase (Fig.\ref{fig:peg_in_hole_approach}(a, b)) and an insertion phase (Fig.\ref{fig:peg_in_hole_approach}(c)).
The search phase has two steps: Learning-based visual servoing (Fig.\ref{fig:peg_in_hole_approach}(a)) and spiral search (Fig.\ref{fig:peg_in_hole_approach}(b)).
The learning-based visual servoing quickly moves a peg closer to the hole, while the spiral search can precisely align the peg and the hole. 

\begin{figure}[!htbp]
    \centering
    \includegraphics[width=0.95\columnwidth]{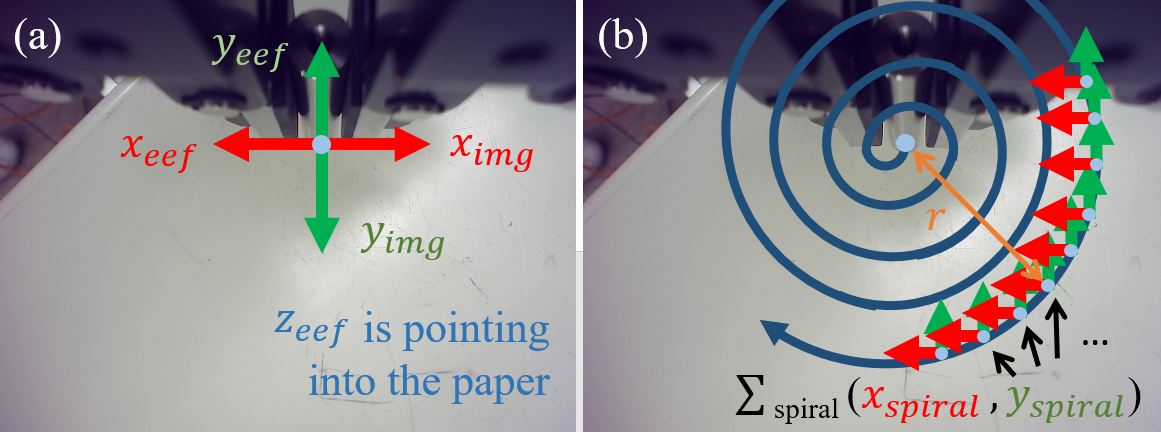}
    \caption{Definition of the coordinate systems $\Sigma_{eef}$, $\Sigma_{img}$,
    and $\Sigma_{spiral}$. (The definition is for the image captured by Camera 1.
    For the other camera discussed later, the coordinate system reverses.)}
    \label{fig:coordinate_system}
\end{figure}

The details of the learning-based visual servoing will be discussed in
Section \ref{visual_alignment}. It is the main contribution of the work.

The spiral search is conducted along the $xy$-plane of the end-effector
coordinate system $\Sigma_{eef}(x_{eef},y_{eef})$, shown in Fig.\ref{fig:coordinate_system}.
The reference coordinate system for spiral search
$\Sigma_{spiral}(x_{spiral},y_{spiral})$ is of the same orientation as $\Sigma_{eef}$
with the origin $r$ away from the initial peg position. The path
for spiral search is given in Eqn.\eqref{eq:spiral_search}.
\begin{equation} \label{eq:spiral_search}  
x_{spiral} = r\cos{\theta}, y_{spiral} = r\sin{\theta}
\end{equation}
where $\theta$ and $r$ start from 0. $\theta$ increases by $\delta\theta$ every timestep, 
while $r$ increases for $\delta$$r$ for every 1 full rotation. 
The robot will move following the discrete spiral path described above, 
and continue until the force at the -$z$ direction of
$\Sigma_{eef}$ is less than $F_{max}$ or $r$ reaches a predefined threshold.

The assembly switches to the insertion phase when the condition for the -$z$ direction 
force is fulfilled. Impedance control is used to perform insertion in the insertion phase.
\subsection{Learning-based visual servoing}\label{visual_alignment}

The learning-based visual servoing uses (1) a deep neural network and multi-view images
to predict the position of the hole and (2) continuous visual servoing to move the peg towards the hole.

\subsubsection{Predicting the position of the hole}\label{hole_detector}
We train a neural network that can map an image $I$ to an output of ($x$,$y$), where
($x$,$y$) indicates the distance between the center of the peg and the hole 
as seen in the image in pixels on coordinate system $\Sigma_{img}$ shown in Fig.\ref{fig:coordinate_system}. 

\begin{figure}[!htbp]
    \centering
    \includegraphics[width=\columnwidth]{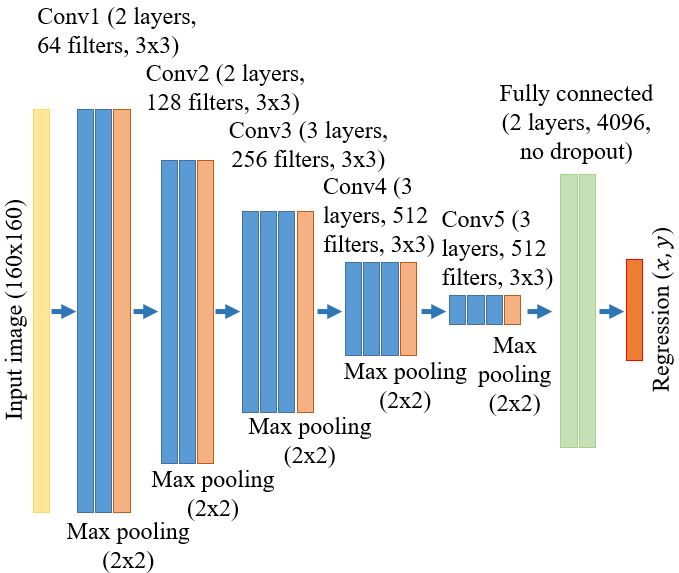}
    \caption{VGG-16 network architecture \cite{simonyan2014very} used in the proposed method.}
    \label{fig:vgg16}
\end{figure}

VGG-16 network is used following the suggestion of \cite{tobin2017domain}. 
The VGG-16 network is adjusted for regression instead of classification.
The input of the network is adjusted to a grayscale image of size 160$\times$160. 
The output is a predicted hole position ($x$,$y$). 
Fig.\ref{fig:vgg16} shows the diagram of the VGG-16 network. 
Following \cite{tobin2017domain},
the dropout components of the network are removed to avoid local minima. 
The network is trained using Adam, with settings following \cite{kingma2014adam}.
Mean squared error (MSE) is selected as the loss function.

\begin{figure}[!htbp]
    \centering
    \includegraphics[width=1.\columnwidth]{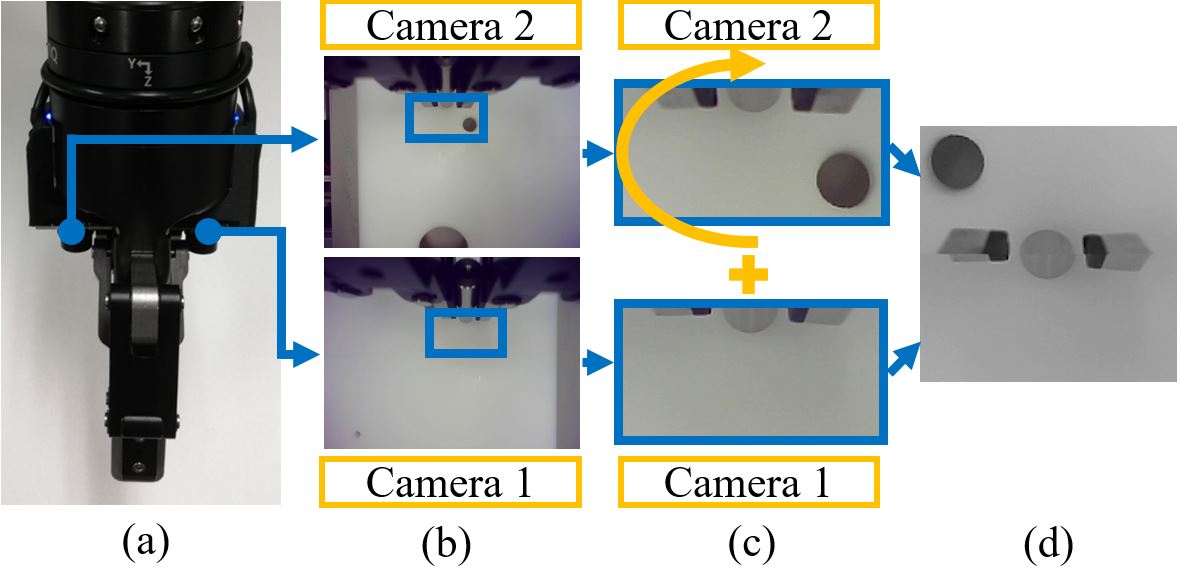}
    \caption{Concatenating two multi-view images into the input image to the VGG-16 network.
    (a) Configurations of the two in-hand cameras. (b) An area around the center of the peg of each image is cropped. 
    (c-d) Concatenate the cropped area into a 160$\times$160 image.}
    \label{fig:real_img_preproc}
\end{figure}

The input image is a concatenation of two images from two in-hand cameras
installed on two sides of a robot hand.
Fig.\ref{fig:real_img_preproc} shows the concatenation. To define the size of the cropped
image, a bounding box of size 160$\times$80 around the center of the peg is predefined,
such that the concatenated image reaches 160$\times$160.

\subsubsection{Iterative visual servoing}
\begin{figure}[!htbp]
    \centering
    \includegraphics[width=0.95\columnwidth]{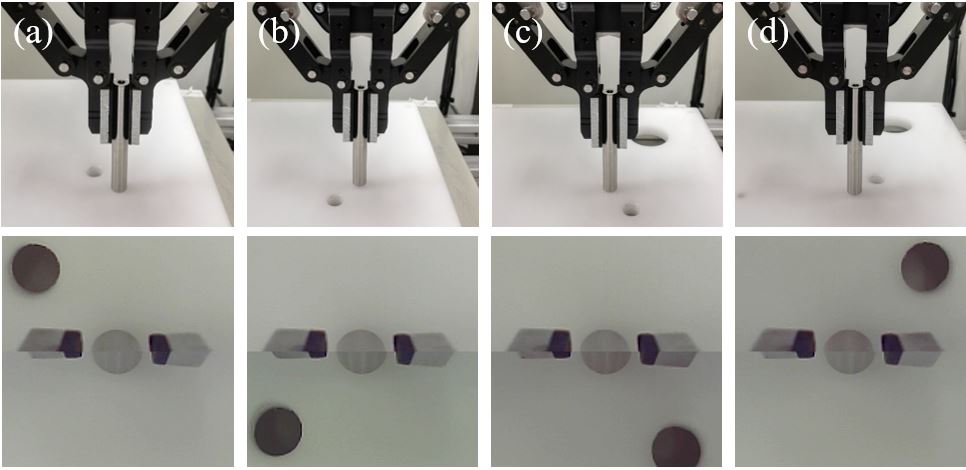}
    \caption{Definition of the quadrants. Top images show how the gripper and 
    the hole are relatively positioned from a front view. The bottom images are 
    the concatenated image from the two cameras. (a) Quadrant "Topleft". 
    (b) Quadrant "Bottomleft". (c) Quadrant "Bottomright". (d) Quadrant "Topright".}
    \label{fig:quadrant_visual_servoing}
\end{figure}

While the network outputs ($x$,$y$) in pixels, the image is not exactly a 2D 
image parallel to the surface of where the hole is. Thus, instead of directly moving
the peg to the predicted position, we use the $\sgn$ function to classify the
outputted values into 4 quadrants, as shown in Fig.\ref{fig:quadrant_visual_servoing},
and iteratively moves the peg towards the quadrants.

Consider the coordinate system $\Sigma_{h}$ which shares the same orientation
as $\Sigma_{eef}$ and has an origin at the center of the hole. 
Assuming at discrete timestep $t$, where $t$ is a non-negative integer that starts from 0, 
the peg is located at ($x_{h}[t]$,$y_{h}[t]$) and the values outputted by the trained 
network is ($x[t]$,$y[t]$). We can move the peg closer with only the quadrant information
using Eqn.\eqref{eq:visual_servoing}.

\begin{equation} \label{eq:visual_servoing}
\begin{bmatrix}
x_{h}[t+1] \\
y_{h}[t+1]
\end{bmatrix}
=\begin{bmatrix}
x_{h}[t] \\
y_{h}[t]
\end{bmatrix}-\lambda[t]\begin{bmatrix}
-\sgn(x[t]) \\
-\sgn(y[t])
\end{bmatrix}
\end{equation}
where $\lambda$ (unit=$mm$/px) is a time dependent coefficient with
decreasing values and converges to 0 along with time. $\lambda$ is defined as:
\begin{equation} \label{eq:visual_servoing_coef}  
\lambda[t]=\frac{A(n-t)}{n}
\end{equation}
where $A$ is the maximum allowable relative moving distance. 
$\lambda[t]$ converges to 0 at time $n$.
By repeating this for $n_{run}$ times, where $n_{run}<n$, 
the peg will get closer to the center of the hole as long as the quadrant the hole
is at relative to the peg can be correctly predicted by the deep neural network.
The method is robust to the prediction errors of the deep neural network since it is not
directly using the predicted numbers.

\subsection{Synthetic Data Generation Method}\label{synthetic_data_gen}
\begin{figure}[!htbp]
  \centering
  \includegraphics[width=1.0\columnwidth]{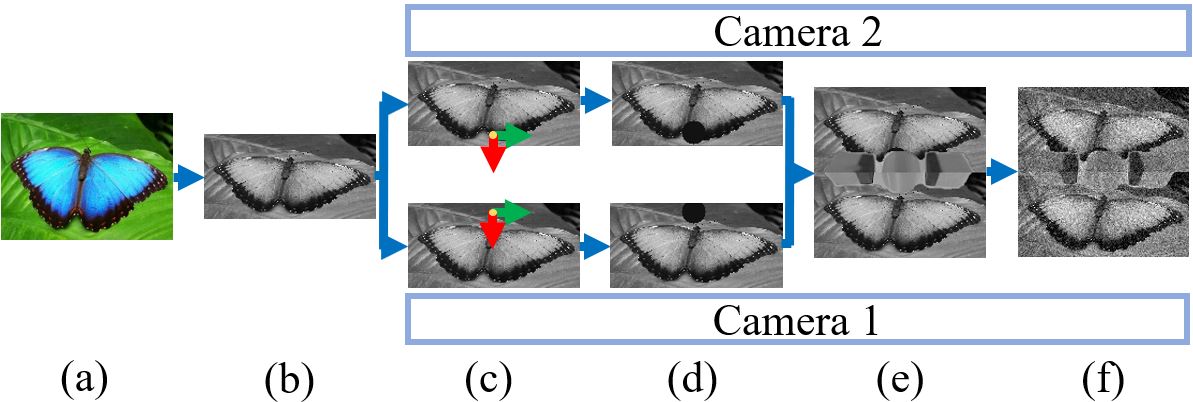}
  \caption{Attaching the gripper template mask to some random images
  with a circle (the hole) to make a synthesized assembly data. The hole
  is added to the background image in (d). The gripper template mask is
  added in (e). Gaussian noises are added in (f).}
  \label{fig:synthetic_data_gen}
\end{figure}

Synthetic data generation is used to get a large amount of training.
The basic idea is to change the background of the cameras with various images.
First, we get a gripper template mask following Fig.\ref{fig:get_template_mask}.
The purpose of having a gripper template mask is to simulate the view of the gripper
in the cameras. Then, the gripper template mask is attached to some random images
with a circle (the hole) to make a synthesized assembly data.
Fig.\ref{fig:synthetic_data_gen} shows attaching process. There are four kinds of 
randomization in the attaching. 
(1) The background of the image is randomized. 
(2) The size of the circle (the hole) is randomized. 
(3) The darkness of the circle (the hole) is randomized.
(4) Gaussian noises are added to randomize the gripper template mask. 
By using random background images captured from the Internet and
likening the hole to a dark-colored circle, a large number of synthetic images
with known labels ($x$,$y$) can be quickly synthesized.

\begin{figure}[!htbp]
    \centering
    \includegraphics[width=0.87\columnwidth]{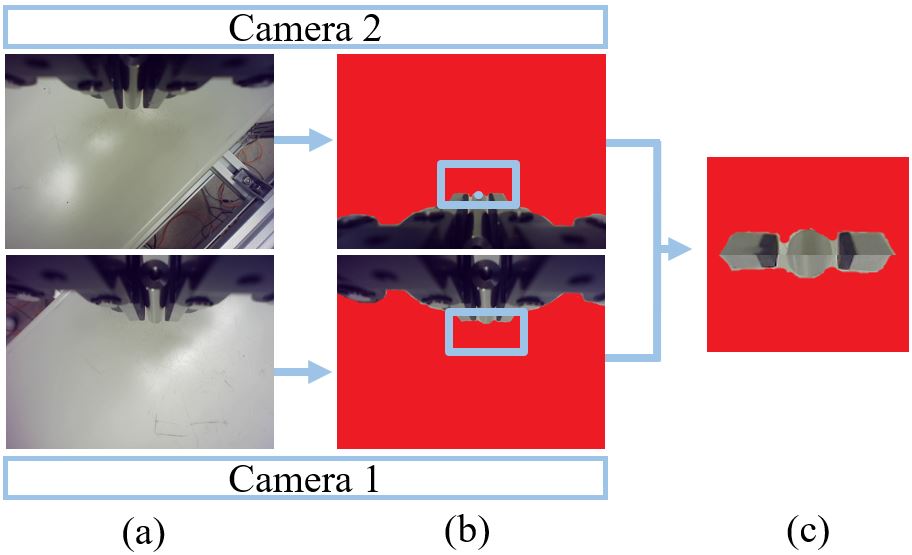}
    \caption{Obtaining the gripper template mask for generating synthetic data.
    (a) The images from the in-hand cameras. (b) Rotate the image from camera 2,
    mark the gripper mask, locate the center of the peg, 
    and define the bounding box on each image. (c) Crop the image according to the bounding box and concatenate.}
    \label{fig:get_template_mask}
\end{figure}

\section{Experiments}\label{experiments}

The experiments section is divided into two parts. In the first part,
we compare and analyze the performance of the neural network under different training data.
In the second part, we analyze the real-world visual servoing and insertions
using the best performing network.

\subsection{Performance of the neural network}

The specification of the computer used for the neural network is
Intel(R) Core(TM) I5-6500 @3.20 GHz, 16GB RAM,
with an Nvidia Geforce GTX 1080 card. 

\subsubsection{Training data}
\begin{figure}[!htbp]
    \centering
    \includegraphics[width=1.\columnwidth]{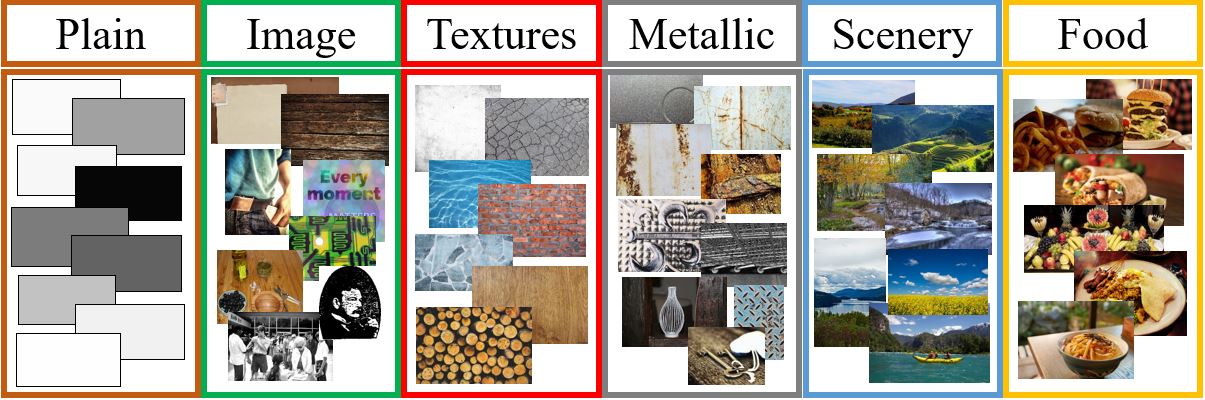}
    \caption{Six categories of randomly collected images are used for synthesizing
    the training data. Images of category "Plain" is generated manually. 
    The rest are downloaded from the Internet using category names as the search keywords.}
    \label{fig:6categories}
\end{figure}

The synthetic training data was generated using the method described in Section \ref{synthetic_data_gen}. 
Six categories of random images were prepared, as shown in Fig.\ref{fig:6categories}.
776 positions (194 positions per quadrant) are evenly sampled in each image to define the position of
a hole. These positions have a maximum of 4 $cm$ uncertainty (the range is [-66 px, 66 px]).
The darkness of a hole was randomized in range [10, 70] (0 is fully black, 255 is fully white).
The diameter of a hole (in pixels) was randomized in range [10 px , 35 px] (around [3 $cm$, 1 $cm$]).
In total, we generated 69,840 synthetic images using each category of random images.
It took an average time of 22 minutes without using the Graphics Processing Unit (GPU).

The details of the synthesis are as follows. Using the six categories of images,
we synthesized 9 sets of training data. They are
``Plain'', ``Image'', ``Textures+Scenery(18)'', ``Textures+Scenery(30)'',
``Textures+Scenery(45)'', ``Textures+Scenery(90)'',
``Textures(45)'', ``Metallic(45)'', ``Scenery(45)''.
For the ``Plain'' training set, the background images were randomly selected 
from the ``Plain'' category, resulting into 69,840 synthetic images 
with background color ranges [0,255].
For the ``Image'' training set, the background images were randomly selected 
from the ``Image'' category where 776 random images searched using the keyword ``Image'' 
were downloaded.
For the ``Textures(45)'', ``Scenery(45)'', and ``Metallic(45)'' training set,
45 images randomly selected from their corresponding categories. 
For the ``Textures+Scenery(45)'' training set, 22 images from the ``Textures'' category and 23 images
from the ``Scenery'' category were randomly selected and 
combined. 
The ``Textures+Scenery(18)'', ``Textures+Scenery(30)'', and ``Textures+Scenery(90)''
training data sets were prepared similarly to "Textures+Scenery(45)",
except that different number of images (9-9, 15-15, and 45-45 respectively) were selected
from the corresponding categories.
Images from the ``Food'' category were not used in the training data. They were prepared for testing.

\subsubsection{Testing data}
Generation of testing data was done similarly, except with 584 random positions
instead of 776. In total, 6 sets of testing data were prepared. They are
``Plain'', ``Light plain'', ``Textures'', ``Metallic'', ``Scenery'', and ``Food''.
The background images of these testing data set were randomly selected from
their corresponding categories. Especially,
for the "Light plain" testing set, 35 different colors of range [125,255] were 
selected instead of [0,255], making it different from the ``Plain'' testing set.

\subsubsection{Training}

Several different VGG-16 networks were trained and compared using the
various training data sets. The names of the networks are the same as the training data sets
to clearly show the correspondence. 
The parameter settings of the VGG-16 neural network was shown in Fig.\ref{fig:vgg16}.
The initial weights were random. The learning rate was set to 1e-5. 
The training data set used for each network was divided by a ratio of 8:2
for training and validation. The epoch was set to 40. Convergence was faster
for less random images (``Plain'', ``Textures'', ``Metallic surface''). The loss at
the end of epoch 40 for these less varied image categories was also smaller,
albeit overfitting existed in all trained networks, similar to 
\cite{tobin2017domain}. Each training time was on average 11 hours.

\subsubsection{Results}

Table \ref{Tab:performance_synthetic_dataset} shows the results of the trained networks and their performance on the
6 sets of testing data.

\begin{table}[!htbp]
\centering
\caption{Performance of the VGG-16 networks}
\label{Tab:performance_synthetic_dataset}
\resizebox{\columnwidth}{!}{%
\begin{threeparttable}
\begin{tabular}{llcccc}
\toprule
Training data set & Testing data set & $MSE_{all}$ & $MSE_{no_outlier}$ & $R_{outlier} $ &$R_{quadrant}$ \\
\midrule
Plain      & Plain & 95.0 & 5.0 & 0.054 & 0.951
\\
Plain      & Light plain & 0.4 & 0.4 & 0.000 & 0.999
\\
Plain      & Textures & 620.1 & 17.2 & 0.368 & 0.721
\\
Plain      & Metallic & 609.0 & 29.2 & 0.383 & 0.715
\\
Plain      & Scenery & 1396.9 & 72.3 & 0.791 & 0.405
\\
Plain      & Food & 1133.5 & 77.5 & 0.744 & 0.479
\\
Image      & Plain & 396.3 & 12.3 & 0.216 & 0.819
\\
Image      & Light plain & 4.2 & 3.8 & 0.001 & 0.992
\\
Image      & Textures & 185.9 & 13.0 & 0.114 & 0.911
\\
Image      & Metallic & 134.4 & 11.7 & 0.072 & 0.935
\\
Image      & Scenery & 119.1 & 13.6 & 0.070 & 0.940
\\
Image      & Food & 101.0 & 13.2 & 0.065 & 0.934
\\
Textures+Scenery(18)      & Plain & 511.9 & 13.6 & 0.282 & 0.755
\\
Textures+Scenery(18)      & Light plain & 2.3 & 2.0 & 0.0004 & 0.995
\\
Textures+Scenery(18)      & Textures & 352.9 & 15.9 & 0.199 & 0.829
\\
Textures+Scenery(18)      & Metallic & 329.6 & 16.2 & 0.158 & 0.854
\\
Textures+Scenery(18)      & Scenery & 288.5 & 19.8 & 0.158 & 0.862
\\
Textures+Scenery(18)      & Food & 322.9 & 25.9 & 0.221 & 0.824
\\
Textures+Scenery(30)      & Plain & 414.0 & 9.7 & 0.244 & 0.791
\\
Textures+Scenery(30)      & Light plain & 0.99 & 0.99 & 0.000 & 0.998
\\
Textures+Scenery(30)      & Textures & 241.1 & 10.2 & 0.140 &0.884
\\
Textures+Scenery(30)      & Metallic & 225.3 & 11.0 & 0.112 & 0.903
\\
Textures+Scenery(30)      & Scenery  & 196.4 & 13.1 & 0.107 &0.909
\\
Textures+Scenery(30)      & Food  & 298.2 & 17.5 & 0.191 & 0.845
\\

Textures+Scenery(45)      & Plain & 399.2 & 13.4 & 0.259 & 0.804
\\
Textures+Scenery(45)      & Light plain & 2.9 & 2.3 & 0.001 & 0.994
\\
Textures+Scenery(45)      & Textures & 378.9 & 14.4 & 0.210 & 0.826
\\
Textures+Scenery(45)      & Metallic & 359.8 & 14.5 & 0.173 & 0.855
\\
Textures+Scenery(45)      & Scenery & 332.1 & 18.9 & 0.176 & 0.849
\\
Textures+Scenery(45)      & Food & 511.0 & 23.0 & 0.285 & 0.772
\\

Textures+Scenery(90)      & Plain & 447.2 & 12.1 & 0.280 & 0.780
\\
Textures+Scenery(90)      & Light plain & 9.3 & 3.9 & 0.006 & 0.984
\\
Textures+Scenery(90)      & Textures& 389.2 & 13.1 & 0.213 &0.835
\\
Textures+Scenery(90)      & Metallic & 341.0 & 13.8 & 0.166 & 0.863
\\
Textures+Scenery(90)      & Scenery & 345.6 & 16.7 & 0.185 & 0.857
\\
Textures+Scenery(90)      & Food & 380.2 & 17.6 & 0.210 & 0.828
\\

Textures(45)      & Plain & 383.4 & 11.0 & 0.234 & 0.807
\\
Textures(45)      & Light plain & 1.9 & 1.9 & 0.000 & 0.992
\\
Textures(45)      & Textures & 315.7 & 12.3 & 0.167 & 0.860
\\
Textures(45)      & Metallic & 310.1 & 14.2 & 0.151 & 0.863
\\
Textures(45)      & Scenery & 369.0 & 18.6 & 0.185 & 0.837
\\
Textures(45)      & Food & 457.3 & 26.9 & 0.287 & 0.781

\\

Metallic(45)      & Plain & 495.8 & 12.5 & 0.299 & 0.766
\\
Metallic(45)      & Light plain & 7.4 & 3.2 & 0.006 & 0.991
\\
Metallic(45)      & Textures & 409.5 & 12.8 & 0.241 & 0.814
\\
Metallic(45)      & Metallic & 350.9 & 15.0 & 0.192 & 0.858
\\
Metallic(45)      & Scenery & 450.7 & 19.0 & 0.244 & 0.808
\\
Metallic(45)      & Food & 514.0 & 22.0 & 0.282 & 0.775
\\

Scenery(45)      & Plain & 500.0 & 14.4 & 0.282 & 0.769
\\
Scenery(45)      & Light plain & 11.7 & 4.6 & 0.007 & 0.992
\\
Scenery(45)      & Textures & 318.7 & 15.3 & 0.203 &0.862
\\
Scenery(45)      & Metallic & 310.4 & 16.0 & 0.174 & 0.877
\\
Scenery(45)      & Scenery & 285.2 & 18.9 & 0.172 &0.890
\\
Scenery(45)      & Food & 395.4 & 21.0 & 0.236 &0.836
\\

\bottomrule
\end{tabular}
\begin{tablenotes}
\item[Meanings of abbreviations] $MSE_{all}$: Mean squared error on all 
images of the testing dataset; $MSE_{no_outlier}$: Mean squared error of images 
of the testing dataset, excluding the outliers; $R_{outlier}$: Ratio of image 
classified as outliers (images with a mean squared error of more than 
200); $R_{quadrant}$: Ratio of images correctly classified into its 
corresponding quadrants.
\end{tablenotes}
\end{threeparttable}
}
\end{table}

The results show that the ${MSE}_{all}$ on most testing data sets are quite large.
The reason is because of the existence of outliers (the images which
predicted outputs are completely off from the true outputs). Without counting
the outliers, the MSE ($MSE_{no\_outlier}$) drops significantly. Thus, to 
minimize the effect of the outliers, the quadrants and 
iterative visual servoing method explained in section \ref{visual_alignment} was adopted.

\begin{figure}[!htbp]
    \centering
    \includegraphics[width=.97\columnwidth]{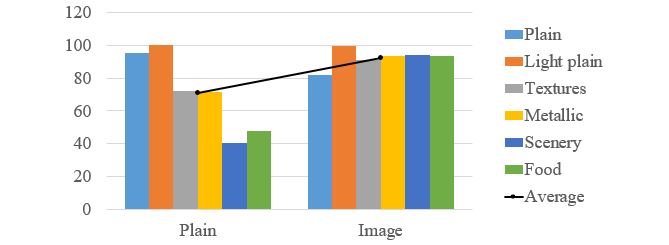}
    \caption{Performance of the networks trained with the ``Plain'' and ``Image'' data set. The vertical axis is
    the $R_{quadrant}$ of Table \ref{Tab:performance_synthetic_dataset} in \%.}
    \label{fig:plain_vs_image}
\end{figure}

Fig.\ref{fig:plain_vs_image} shows how the use of images instead of plain
backgrounds improves the network's robustness. There
is an increase in average performance ($R_{quadrant}$ of Table \ref{Tab:performance_synthetic_dataset}) on all testing data sets. 
The network trained with the ``Plain'' data set performs better on 
the ``Plain'' testing data set due to overfitting, especially in cases where 
backgrounds have similar darkness to the hole.

\begin{figure}[!htbp]
    \centering
    \includegraphics[width=\columnwidth]{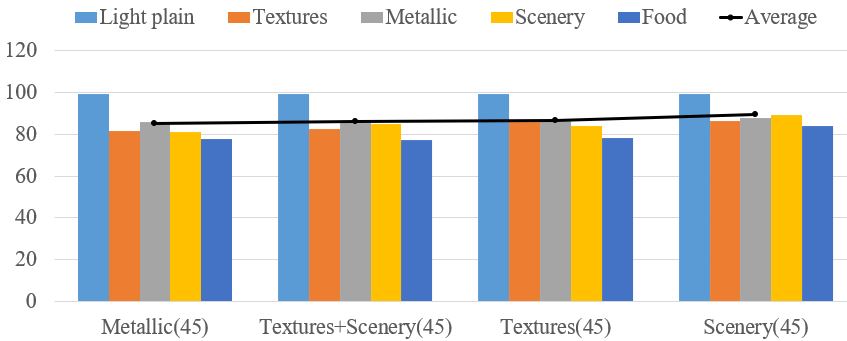}
    \caption{Performance the networks trained with ``Metallic(45)'', 
    ``Textures+Scenery(45)'', ``Textures(45)'', and ``Scenery(45)'' training sets.
    The vertical axis is
    the $R_{quadrant}$ of Table \ref{Tab:performance_synthetic_dataset} in \%.}
    \label{fig:45backgrounds}
\end{figure}

Fig.\ref{fig:45backgrounds} shows the $R_{quadrant}$ of the
``Metallic(45)'', ``Textures+Scenery(45)'', ``Textures(45)'', and ``Scenery(45)'' training sets.
The comparison indicates that: (1) With exception on the ``Light plain'', the networks trained with a certain data set
generally perform better on similar test sets. 
(2) Networks trained on certain data sets cannot perform 
as well on a background with high randomness like the ``Food'' test set.
Thus, training with images of a certain data set can improve the network's performance 
on similar test sets. On the other hand, for categories with less variety like ``Metallic'' or 
``Textures'' (less than ``Food''), improvements on the network performance can be 
obtained by training on data of higher variety, like ``Scenery''.
This is from the observation that the network trained with ``Scenery'' performs similarly to 
the one trained with ``Textures''  on the ``Textures'' test set, and performs better 
compared to the networks trained with ``Metallic'' on the ``Metallic'' test set.

\begin{figure}[!htbp]
    \centering
    \includegraphics[width=\columnwidth]{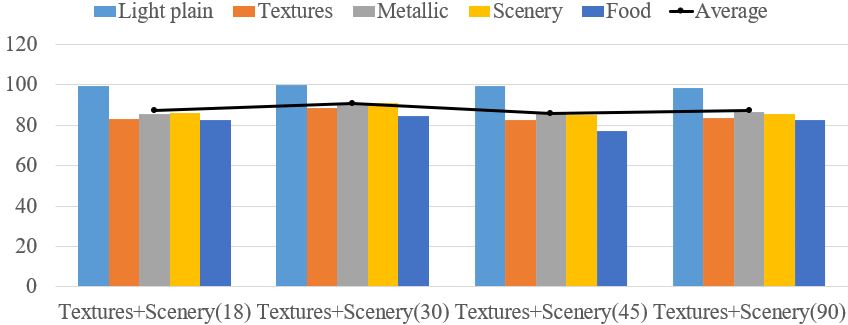}
    \caption{Performance of the networks trained with ``Textures+Scenery'' data sets
    synthesized with different numbers of similar background image. The vertical axis is
    the $R_{quadrant}$ of Table \ref{Tab:performance_synthetic_dataset} in \%.}
    \label{fig:different_number_background}
\end{figure}

Fig.\ref{fig:different_number_background} compares the performance of networks trained
data sets synthesized with different numbers of similar background images. 
The results imply that the number is not the only crucial parameter. Other forms of 
randomization including hole sizes and hole darkness also play 
important roles in the synthesis of the training data.

\subsection{Real-world Experiments}\label{exp_real_world}
\begin{figure}[!htbp]
    \centering
    \includegraphics[width=.97\columnwidth]{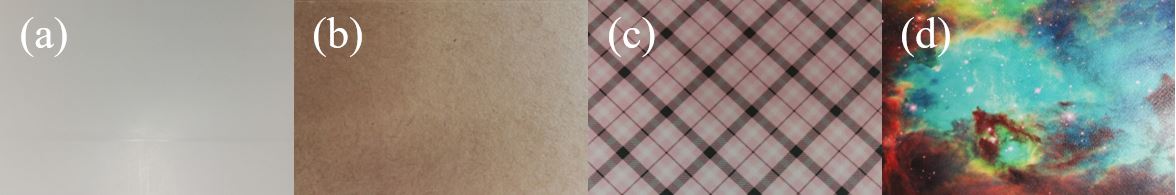}
    \caption{The four surfaces used in real-world experiments. (a) White. (b) Brown. (c) Pink. (d) Sky.}
    \label{fig:surface_real_world_exp}
\end{figure}

We performed real-world experiments using the networks trained with the ``Image'' data set
and the ``Plain'' data set, for they have the best and worst performance.
Four different surfaces, as is shown in Fig.\ref{fig:surface_real_world_exp}, were used in the experiments.
A success execution is judged to be when the peg is inserted within 90 sec. 

\begin{figure}[!htbp]
    \centering
    \includegraphics[width=.95\columnwidth]{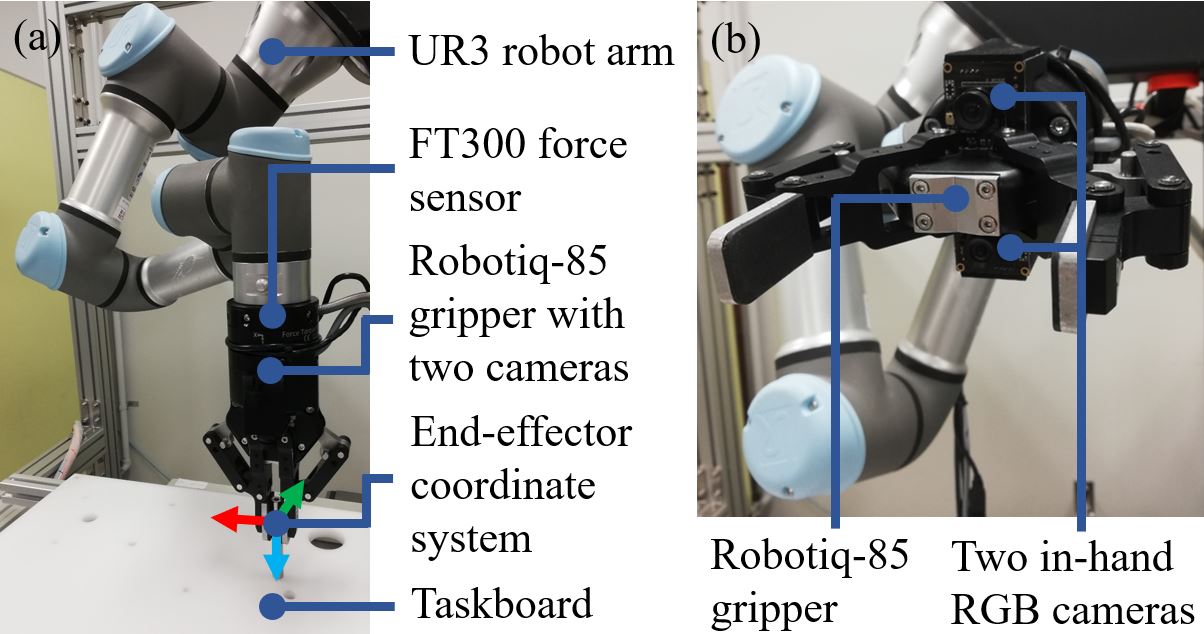}
    \caption{The experimental setup. (a) The overall view on the experimental 
    setup. (b) A close-up view of the Robotiq-85 gripper with the two in-hand 
    RGB cameras.}
    \label{fig:experiment_setup}
\end{figure}

The robot we used to do real-world experiments is a UR3 robot with a FT300 force sensor
and a Roboti-85 gripper. Two in-hand cameras were used to collect multi-view images.
Fig.\ref{fig:experiment_setup} shows the experimental setup. 
The robot was made to insert a 75$\times$10 $mm$ peg into a hole.
The taskboard in the left of Fig.\ref{fig:experiment_setup} is the base with the hole to be inserted. 
The lenience of the hole for the experiment is 0.4$mm$. 
The specification of the computer used was the same as the one used to train the deep neural network. 

The parameters for the spiral search were $r_{init}$=0.3$mm$, $r_{max}$=7.0$mm$, 
$\delta\theta$=12.5$^\circ$, $\delta$$r$=0.3$mm$, and $F_{max}$=20$N$. 
The parameters for the iterative visual servoing were $A$=10$mm$, $n$=10, $n_{run}$=5.
The parameters of the impedance control were $c$ = [50,50,50,1,1,1] for the damper
and $k$ = [100,100,100,100,100,100] for the spring.

\begin{figure}[!htbp]
    \centering
    \includegraphics[width=1.\columnwidth]{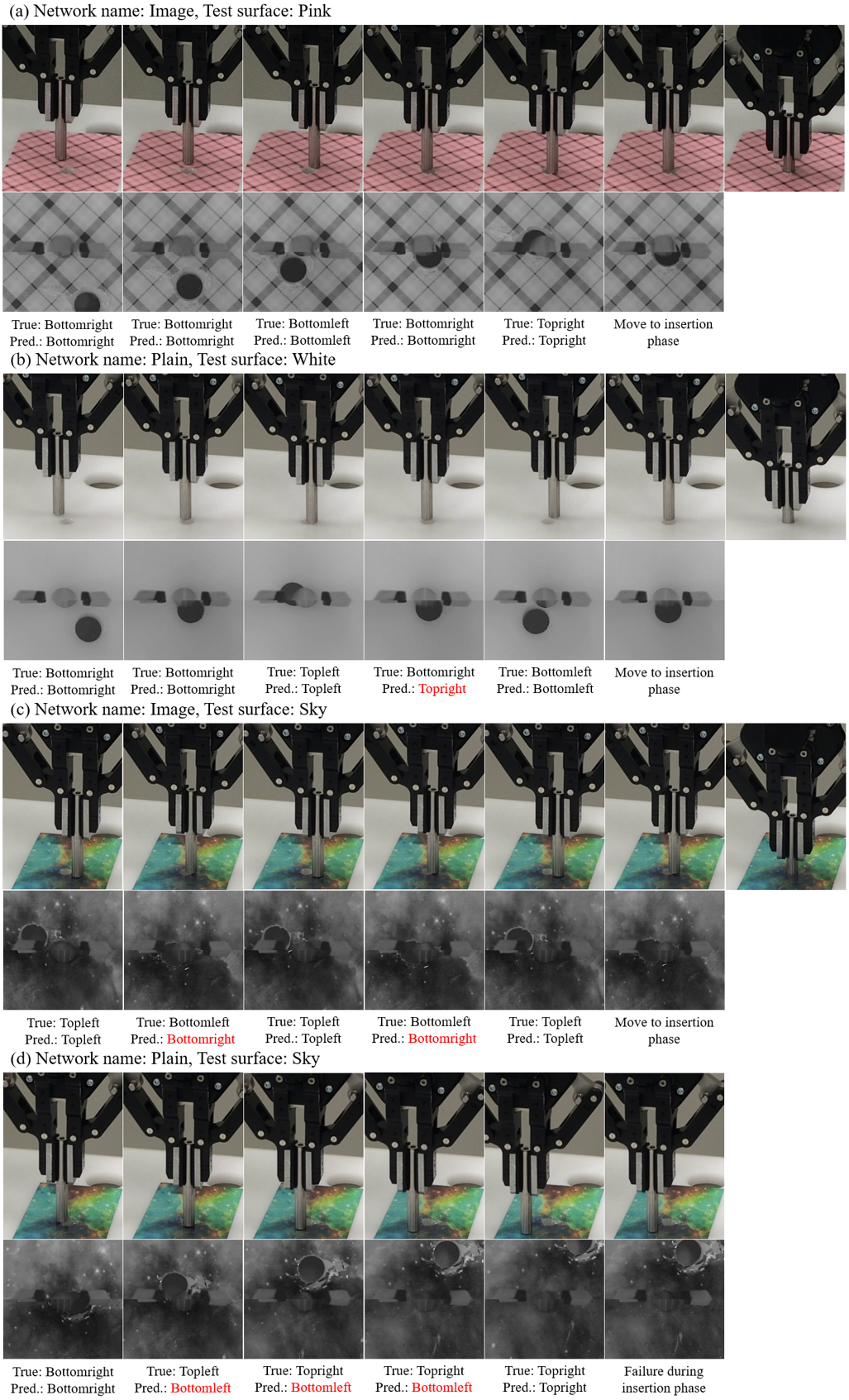}
    \caption{Some snapshots of the real-world executions from the results
    shown in Table \ref{Tab:performance_real_world}. (a) A successful execution using 
    the network trained with ``Image'' and tested on the ``Pink'' surface ($\oplus$ category). 
    (b) A successful execution (with overshoots) using the network trained with ``Plain''
    and tested on the ``White'' surface ($\ominus$ category). (c) A failure case using the network trained with
    ``Image'' and tested on the ``Sky'' surface ($\bigtriangleup$ category). (d)
    A failure case using the network trained with ``Plain'' and tested on the ``Sky'' surface ($\times$ category).}
    \label{fig:real_exp_image}
\end{figure}

Table \ref{Tab:performance_real_world} shows the success rates of the real-world experiments.
Ten times of trial are performed for each training set and testing surface combination. 
The network trained with the ``Image'' data set can successfully ignore 
the variations of the ``White'', ``Brown'', and ``Pink'' surfaces and 
correctly predict the correct quadrants. Fig.\ref{fig:real_exp_image}(a)
shows an example of the success sequence on the ``Pink'' surface.
The network trained with ``Plain'' can also correctly 
predict the quadrant and successfully performs insertion on the ``White'' and ``Brown'' surfaces. 
Although a few overshoots were spotted, the quadrant-based visual servoing 
compensated them after several iterations. An example is shown in Fig.\ref{fig:real_exp_image}(b).

\begin{table}[!htbp]
\tiny
\centering
\caption{Performance of the real-world experiments}
\label{Tab:performance_real_world}
\resizebox{\columnwidth}{!}{%
\begin{threeparttable}
\begin{tabular}{lcccc}
\toprule
Network name & $R_{white}$ & $R_{brown}$ & $R_{pink}$ & $R_{sky}$ \\
\midrule
Image & 10/10 ($\oplus$) & 10/10 ($\oplus$) & 10/10 ($\oplus$) & 3/10 ($\bigtriangleup$)
\\
Plain & 10/10 ($\ominus$) & 10/10 ($\ominus$) & 4/10 ($\bigtriangleup$) & 0/10 ($\times$)
\\

\bottomrule
\end{tabular}
\begin{tablenotes}
\item[Meanings of abbreviations] $R_{white}$: Ratio of successful insertions on the 
``White'' surface; $R_{brown}$: Ratio of successful insertions on the ``Brown'' surface; 
$R_{pink}$: Ratio of successful insertions on the ``Pink'' surface; $R_{sky}$: Ratio
of successful insertions on the ``Sky'' surface; $\oplus$: The deep network 
gives output of the correct quadrant; $\ominus$: The deep network successfully predicts
the correct quadrant; $\bigtriangleup$: The deep network sometimes 
successfully predicts the correct quadrant; $\times$: The deep network rarely 
successfully predicts the correct quadrant.
\end{tablenotes}
\end{threeparttable}
}
\end{table}

\begin{table}[!htbp]
\tiny
\centering
\caption{Time cost under different initial position errors}
\label{Tab:real_world_time}
\resizebox{0.8\columnwidth}{!}{%
\begin{threeparttable}
\begin{tabular}{lcccc}
\toprule
 $D_{euclidean}$      & $t_{with-visual}$      & $t_{without-visual}$ 
\\
\midrule
23.8     & 30.4 & $>$90.0 
\\
15.0    & 33.4 & $>$90.0
\\
12.0     & 41.3 & $>$90.0
\\
11.7     & 47.2 & $>$90.0
\\
10.0    & 70.1 & $>$90.0
\\
12.5    & 39.0 & $>$90.0
\\
11.0    & 63.4 & $>$90.0
\\
14.1    & 22.8 & $>$90.0
\\
4.0    & 56.1 & 25.5
\\
13.6    & 43.4 & $>$90.0
\\
\bottomrule
\end{tabular}
\begin{tablenotes}
\item[Meanings of abbreviations] $D_{h,euclidean}$: Initial distance from the center of the peg 
o the center of the hole in $mm$; $t_{with-visual}$: Time taken from the start of 
the search phase until the end of the insertion using the proposed method in sec. (network trained using ``Image'');
$t_{with-visual}$: Time taken from the start of the search phase until the end of insertion with a simple spiral search in sec.
\end{tablenotes}
\end{threeparttable}
}
\end{table}

The failure appears with the ``Sky'' surface for both networks and the ``Pink'' 
surface for the network trained with the ``Plain'' data set. 
For the network trained with the ``Image'' data set and examined using the ``Sky'' surface
and the network trained with the ``Plain'' data set and examined using the ``Pink'' surface,
errors sometimes occur, resulting into 3/10 and 4/10 success rates in Table \ref{Tab:performance_real_world} respectively.
Fig.\ref{fig:real_exp_image}(c)
shows an example of the failure sequence.
For the network trained with the ``Plain'' data set and examined using the ``Sky'' surface,
the success rate is 0/10. Fig.\ref{fig:real_exp_image}(d) 
is an example of the failure.

Table \ref{Tab:real_world_time} compares the time cost of 10 executions on the ``White'' surface
with and without the learning-based visual servoing. The initial positions errors were randomly set
(as is shown in the first column of Table \ref{Tab:real_world_time}).
When the initial position error is large, 
the robot system could finish an insertion in less than 70 sec. with the learning-based visual servoing.
In contrast, the execution costs larger than 90 sec. or fails. 
An exception is $D_{euclidean}$=4.0$mm$. In this case, the initial position error is small. A simple
spiral search could quickly find the hole. The result demonstrates that the proposed method
improves search efficiency when
the position of the hole has large uncertainty.

\section{Conclusions}\label{conclusion}
In conclusion, we developed a learning-based visual servoing method to quickly 
insert pegs into uncertain holes. The method used a deep neural network trained on synthetic data
to predict the quadrant of a hole, and used iterative visual servoing to move the peg towards the hole
step by step. The synthetic data was generated by cutting and 
pasting gripper template masks on random images, which allowed extremely fast 
synthetic data generation. Performance of different training data sets was compared.
Training with images from categories of higher variety can lead to better 
performance, even for testing with images from categories of less variety. 
Real-world experiments showed that the proposed method is robust to various surface backgrounds. 
The system is generally faster compared to a peg-in-hole
assembly using only a spiral search, unless the initial error is very small.


\bibliographystyle{IEEEtran}
\bibliography{paperJoshua}

\end{document}